\documentclass[runningheads]{llncs}
\usepackage[T1]{fontenc}
\usepackage{makecell}
\usepackage{float}

\usepackage{graphicx}
\usepackage{amsmath}
\usepackage{amsfonts}
\usepackage{array}
\usepackage{ragged2e}
\usepackage{placeins} 

\begin{document}
\titlerunning{SafeML for Runtime Safety of ML Systems}

\title{Incorporating Failure of Machine Learning in Dynamic Probabilistic Safety Assurance}
%
%
\author{Razieh Arshadizadeh\inst{1} 
\and
Mahmoud Asgari\inst{1}\and 
Zeinab Khosravi\and
Yiannis Papadopoulos\inst{1}\and
Koorosh Aslansefat\inst{1}}
\authorrunning{F. Author et al.}
%
\institute{School of Computer Science, University of Hull, Hull HU6 7RX, UK }

\maketitle              
\begin{abstract}

Machine Learning (ML) models are increasingly integrated into safety-critical systems, such as autonomous vehicle platooning, to enable real-time decision-making. However, their inherent imperfection introduces a new class of failure: reasoning failures often triggered by distributional shifts between operational and training data. Traditional safety assessment methods, which rely on design artefacts or code, are ill-suited for ML components that learn behaviour from data. SafeML was recently proposed to dynamically detect such shifts and assign confidence levels to the reasoning of ML-based components. Building on this, we introduce a probabilistic safety assurance framework that integrates SafeML with Bayesian Networks (BNs) to model ML failures as part of a broader causal safety analysis. This allows for dynamic safety evaluation and system adaptation under uncertainty. We demonstrate the approach on an simulated automotive platooning system with traffic sign recognition. The findings highlight the potential broader benefits of explicitly modelling ML failures in safety assessment.

\end{abstract}
\section{Introduction}
\label{intro}
The growing demand for safe and efficient Autonomous Vehicle (AV) systems has spurred interest in platooning, where multiple AVs follow a lead vehicle at a defined inter-vehicle distance. Platooning aims to improve traffic flow, reduce fuel consumption, and enhance road safety by minimizing human errors~\cite{fagnant2015preparing,gautam2021runtime,reich2016systematic,tsugawa2016review}. Typically,  a human-driven lead vehicle is followed by one or more autonomous vehicles that adjust their speeds to maintain safe gaps. This coordination enhances both safety and operational efficiency by reducing reaction times and synchronized movements~\cite{kabir2019runtime,muller2016safety,schneider2013conditional}. However, the safety and effectiveness of platooning systems depend on their ability to adapt to changing road and environmental conditions. Traditional frameworks often rely on static parameters, e.g., fixed speed limits, that may be insufficient in dynamic road contexts.  Unpredictable weather changes or new traffic signs can render such static speed limits inadequate, highlighting the need for real-time adaptation~\cite{gautam2021runtime}.

Recent studies address this through Dynamic Safety Contracts (DSCs) and Conditional Safety Certificates (ConSerts)~\cite{schneider2013conditional}. Despite proving effectiveness by enabling AVs to adapt their behaviours, e.g., to dynamically maintain safety, a significant limitation of DSCs and ConSerts lies in their reliance on binary conditional variables, which may not fully capture the probabilistic nature of real-world traffic scenarios~\cite{gautam2021runtime,kabir2019runtime,muller2016safety,reich2016systematic}. To improve reasoning under uncertainty, BNs have been introduced as a flexible framework for safety assessment in AVs~\cite{kabir2019runtime}. BNs support probabilistic reasoning, integrating diverse factors into cohesive safety evaluation. The use of BNs has demonstrated effectiveness in managing uncertainty in AV systems, particularly for vehicle platooning applications where real-time decision-making is required to ensure safe inter-vehicle distances. Yet, the framework developed in \cite{kabir2019runtime} employs a fixed speed limit, which limits its applicability to dynamically changing traffic scenarios~\cite{gautam2021runtime,kabir2019runtime,reich2016systematic}.

Meanwhile, ML models, particularly Convolutional Neural Networks (CNNs), enable AVs to interpret road scenes, including traffic signs, for adaptive decision making. CNNs excel at identifying visual patterns, making them well-suited for traffic sign recognition. However, they are sensitive to uncertainties, such as sensor noise,  lighting variations, and distributional shifts between training and operational data. These factors can undermine prediction reliability, posing safety risks if left  unmanaged~\cite{mcallister2017concrete}. To mitigate this, techniques like as Probabilistic CNNs (PCNNs) have been developed that can provide uncertainty in their predictions, providing a measure of confidence~\cite{mcallister2017concrete}. Another approach, SafeML, evaluates prediction reliability by measuring distributional shifts of incoming operational data from the training data. When a shift is detected, predictions can be flagged as unreliable, prompting safety-preserving responses like speed reductions~\cite{aslansefat2020safeml}.

In this work, we propose a novel framework that integrates BNs with SafeML to potentially enhance the safety of platooning by dynamically adapting speed limits in response to real-time road signs and environmental conditions. By fusing information from both the ML model and SafeML, the BN enables AVs to maintain safety under uncertainty. The core innovation lies in the explicit modeling of ML failures within a dynamic runtime safety assurance framework. This approach is evaluated in a simulated platooning environment, focusing on its potential to improve safety rather than on real-time performance. While the framework is demonstrated in the context of AV platooning, it is broadly applicable to intelligent systems requiring dynamic, uncertainty-aware safety assessment, offering a flexible and extensible solution for runtime assurance across domains.

\section{Related Works}

\subsection{SafeML}
SafeML employs statistical tools, such as ECDF-based statistical distance measures, to monitor classifier performance at runtime. The method quantifies distributional shifts in upcoming data and estimates potential performance degradation accordingly. It also incorporates a human-in-the-loop mechanism to improve safety under conditions such as concept drift. The concept was introduced by Aslansefat et al. \cite{aslansefat2020safeml}.  The initial idea was extended for image classification in \cite{aslansefat2021toward}. The study introduced bootstrapping-based p-value calculation to validate distances and increase accuracy. SafeML has also been extended for time-series prediction and regression tasks \cite{akram2022stadre}. The issue of defining a well-tuned appropriate threshold of unacceptable drift was investigated in  \cite{farhad2022keep}. The study proposed a method for automatic tuning that optimises performance. 

SafeML has shown promising results in numerous applications such as security intrusion detection \cite{aslansefat2020safeml}, autonomous driving system \cite{aslansefat2021toward}, safety zone estimation and object detection in robotics \cite{cho2022online,aslansefat2022safedrones}, and offshore wind turbine blade inspection using UAVs \cite{kabir2022combining}. H. Farhad et al. \cite{farhad2023scope} proposed a model-specific version of SafeML in which the last layer of a deep neural network and the latent features of the networks have been used to drive statistical distance measures.  The approach has also been extended for Machine Learning explainability. A method called SMILE (Statistical Model-agnostic Interpretability with Local Explanations) \cite{aslansefat2023explaining} uses SafeML statistical distance measures to provide explanations for specific classification in relevant tasks.  SafeML has recently  been cited in the German Industry Standard for Machine Learning Uncertainty Quantification (DIN SPEC 92005) \cite{DIN92005} .

\subsection{Runtime BN and Safety Models}

Bayesian networks (BNs) are probabilistic graphical models with a flexible architecture, capable of reasoning under uncertainty. They provide a global assessment of various dependability properties, such as reliability and availability, by aggregating local information from different sources. Structurally, BNs represent relationships among random variables using a directed acyclic graph (DAG), where nodes represent variables and edges represent conditional dependencies.  If an arc points from node X to node Y, then X is the parent of Y and exerts a direct (deterministic or probabilistic) influence on it \cite{kabir2018dynamic}.  This structure enables both causal reasoning and probabilistic inference. 

In recent years, BNs have gained widespread adoption in dependability engineering, particularly for safety and reliability assessments \cite{kabir2019applications}. Their utility in aggregating and propagating uncertain information makes them well-suited for complex dynamic systems such as AVs. Several studies have explored runtime assurance and adaptability in AV platooning. Early projects such as  PATH and SARTRE pioneered adaptive cruise control, and intervehicle communication with a focus on efficiency and safety of platoons  \cite{kabir2019runtime}.  Müller et al. \cite{muller2016safety} proposed an application of Dynamic Safety Contracts (DSC) to platooning. In this approach, they provided modular runtime checks for qualitative and quantitative safety conditions and highlighted the need for adaptive and situation-aware decision-making in dynamic situations. Schneider and Trapp \cite{schneider2013conditional} proposed conditional safety certificates (ConSerts) for open adaptive systems, delivering runtime safety guarantees via condition-based contracts.

Among more recent advancements, Kabir \textit{et al.} employed BNs to manage uncertainty in AV systems, notably demonstrating their value in safe and adaptive AV platooning \cite{kabir2019runtime}.  However, their framework assumes a fixed speed limit and does not support complex perceptual tasks such as traffic sign recognition.  Moreover, it does not address the uncertainty introduced by ML models. In response, Gautam et al. proposed a hybrid decision-making framework combining probabilistic CNNs (PCNN) and BNs. Their approach supports real-time decision-making under uncertainty, particularly for traffic sign recognition in platooning scenarios \cite{gautam2021runtime}. Nonetheless, their proposal neither explicitly accounts for the imperfect reliability of ML models nor does it address out-of-distribution (OOD) data. 

In our work, we present a new framework that incorporates SafeML as a real-time evaluator within a BN framework. SafeML quantitatively assesses ML uncertainty through statistical measures and feeds this information into the BN to support dynamic reliability modeling. This integration enables the system to adapt to changing data distributions, improving real-time safety assurance. By treating ML model reliability as a first-class input to decision making, we bridge the gap between probabilistic reasoning and dynamic learning-based perception in AV platooning.

\section{Method}

\subsection{Bayesian Risk Assessment Loop}
\begin{figure}
  \centering
  \includegraphics[width=0.75\linewidth, trim=20 10 20 10, clip]{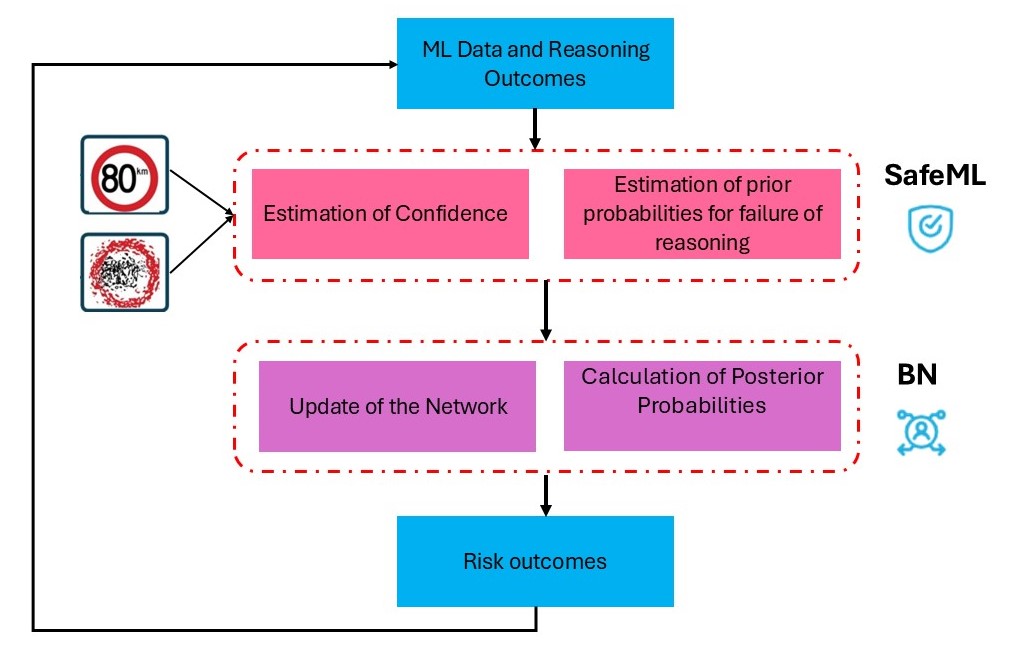}
      \vspace{-1.5em}

  \caption{A cyclic integration of SafeML with a Bayesian Network for confidence estimation, risk assessment, and proactive mitigation.}

  \label{fig:loop}
\end{figure}

The central mechanism of our approach to dynamic risk assessment is a BN that is continuously updated with information about the system's state. To incorporate assessments of ML reasoning failures, we propose a mechanism for connecting SafeML monitors to the BN as illustrated in (see Fig.~\ref{fig:loop}). 

The process begins with the acquisition of ML-generated data and reasoning outcomes, including classification results and contextual input signals. These outputs are passed into the SafeML module, which performs two key functions: 

(i) estimating confidence by measuring  the similarity between distributions of operational  inputs and training data, and

(ii) deriving prior probabilities that reflect the likelihood of reasoning failure,  encapsulating the statistical trustworthiness of the ML model’s predictions.

These  prior probabilities are then fed to the BN module, which carries out  two main inference tasks: 

(a) updating the probabilistic graphical model with real-time evidence, and 

(b) computing posterior distributions over safety-relevant system states.

The final output of this pipeline is a quantified risk outcome that reflects the system's operational safety status. This output can be used not only for decision support but also as feedback input for the next processing cycle that uses fresh ML and other system data. The result is a closed-loop safety monitoring mechanism that adapts continuously to evolving operational conditions, providing robustness against both distributional shifts and reasoning anomalies.


    \vspace{-1em}

\subsection{SafeML-Augmented CNN for Traffic Sign Recognition}
\begin{figure}[!htbp]
    \centering
    \includegraphics[ width=\linewidth, trim=20 10 20 20, clip]{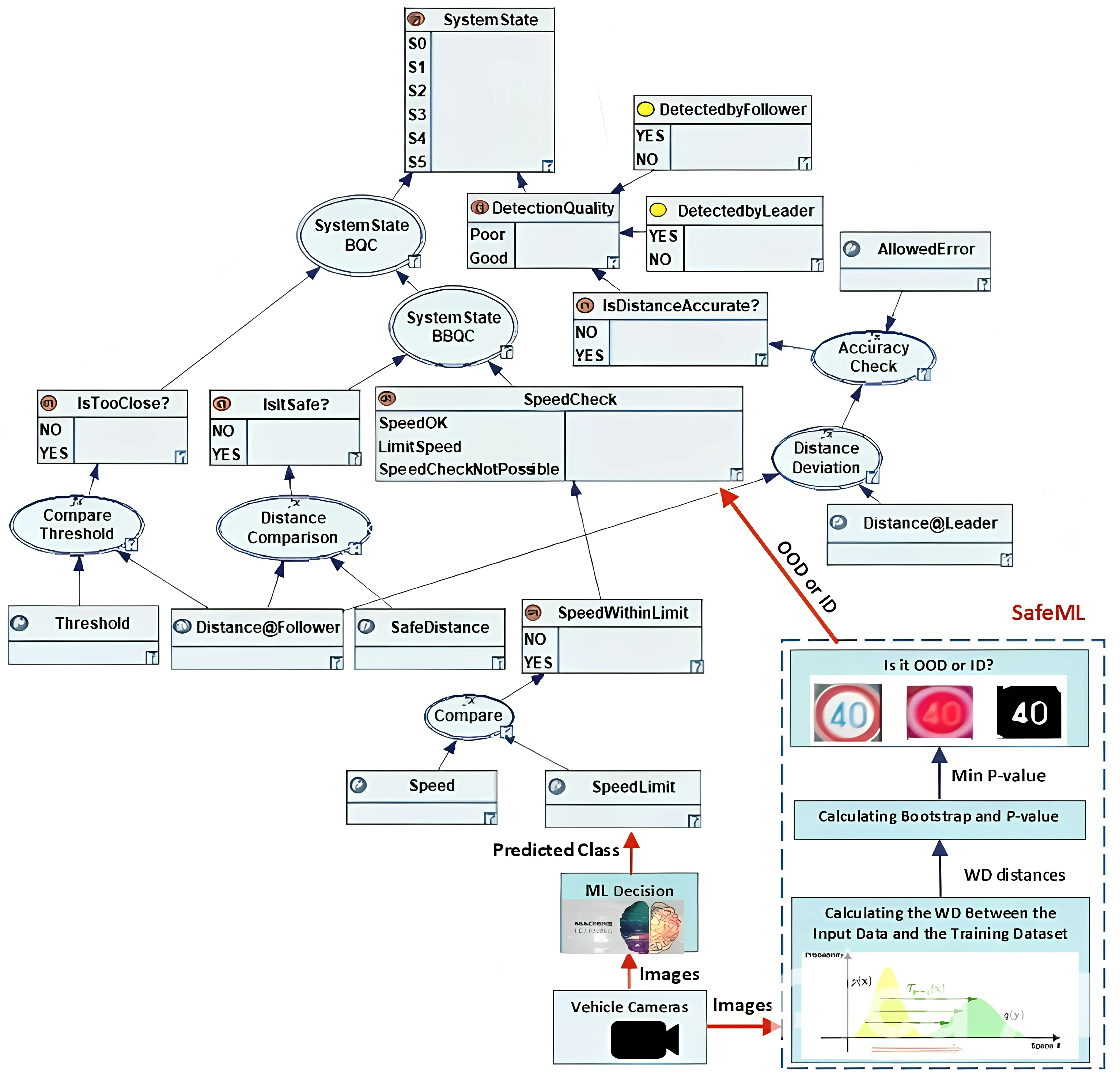}
        \vspace{-2em}

    \caption{Bayesian network integrating ML and SafeML for runtime safety assurance.}
    \vspace{-1em}

    \label{fig:safeml}
\end{figure}

 Figure~\ref{fig:safeml} provides a detailed view of the proposed dynamic safety assurance framework in the context of an AV platooning scenario, illustrating the integration of ML, statistical validation, and probabilistic reasoning with the BN providing dynamic risk estimation.  Our ML component for traffic sign recognition employs a CNN architecture trained on the German Traffic Sign Recognition Benchmark (GTSRB) dataset. The model's architecture involves sequential convolutional layers, max-pooling, dropout regularization for generalization, and fully connected layers to classify traffic signs effectively.
 
 The process begins with raw visual inputs from vehicle-mounted cameras, which are processed by a trained ML classifier to generate traffic sign predictions. These predictions are simultaneously passed to a SafeML module, which evaluates whether the input sample lies within the training distribution (in-distribution, ID) or deviates from it (out-of-distribution, OOD). This evaluation uses the Wasserstein distance between input and training data, followed by a bootstrap-based p-value calculation. SafeML 
 outputs a binary reliability signal (ID or OOD), indicating the trustworthiness of the ML decision. Both the predicted class and the SafeML reliability signal are provided as evidence to the BN, alongside contextual variables such as inter-vehicle distance, speed compliance, sensor accuracy, and system-specific thresholds. The BN captures semantic relationships between these variables, allowing the inference engine to compute posterior distributions over key safety-relevant states (e.g., SystemState, SystemStateBQC, DetectionQuality). Decision nodes such as `SpeedWithinLimit`, `IsItSafe`, and `DistanceComparison` evaluate operational conditions in real time.

The BN output allows the system to assess the overall safety status under uncertainty, differentiate between root causes of potential failure (e.g., ML misclassification vs. sensor error), and issue targeted mitigation actions. Thus, safety-critical decisions are not made solely on raw ML predictions but are contextually informed, statistically validated, and probabilistically reasoned.
\vspace{-1.5em}

\vspace{-1.5em}

\subsection{Mathematical Formulation}

This subsection presents the mathematical background supporting our proposed SafeML-enhanced runtime assurance framework.

\subsubsection{Wasserstein Distance}
The Wasserstein distance between two probability distributions \( P \) and \( Q \) is defined as~\cite{villani2008optimal}:
\begin{equation}
W(P, Q) = \inf_{\gamma \in \Gamma(P, Q)} \int_{\mathbb{R}^n \times \mathbb{R}^n} \|x - y\| \, d\gamma(x, y)
\end{equation}
where \( \Gamma(P, Q) \) denotes the set of all joint distributions with marginals \( P \) and \( Q \), and \( x \), \( y \in \mathbb{R}^n \) represent samples drawn from these distributions.



    \subsubsection{Bootstrapped p-value}
    To statistically assess whether a test input belongs to the same distribution as the training set, we compute a p-value using bootstrap resampling:
    \vspace{-0.5em}
    \begin{equation}
    \text{p-value} = \frac{1}{B} \sum_{b=1}^{B} \mathbf{1} \left( W(X_b^*, X_{\text{train}}) \geq W(\hat{X}, X_{\text{train}}) \right)
    \end{equation}

\subsubsection{Safety Check}

The final decision about the reliability of ML classification is based on the minimum p-value across the three RGB channels:

\begin{equation}
\text{Unreliable} = 
\begin{cases}
0 & \text{if } \min\left(\text{pval}_R, \text{pval}_G, \text{pval}_B\right) > 0.01 \\
1 & \text{[[otherwise]]}
\end{cases}
\end{equation}

\noindent
In other words, the input is flagged as \textit{unreliable} only if all three colour channels yield a statistically significant deviation (p-value > 0.01). This conservative criterion ensures that any statistically significant deviation in a single channel is sufficient to reject the in-distribution assumption.
\vspace{-0.5em}

\subsubsection{Bayesian Integration of ML and SafeML Outputs}

Each system component, ML predictions, SafeML outputs, and contextual signals, is encoded as a node in the main decision-making model, i.e., the BN. These nodes, modeled as random variables, are interconnected in the BN with edges showing interdependencies.
The ML classifier outputs a predicted label \texttt{MLDecision}, while SafeML generates a reliability indicator (\texttt{SafeML\_Status,}i.e., ID or OOD). These two variables condition downstream nodes representing safety-relevant aspects such as the speed compliance, intervehicle distance, and ultimately the system's operational state.

The conditional probability distribution of a node \( V_i \) given its parent nodes is denoted as~\cite{villani2008optimal}:
\begin{equation}
 \Pr\{V_i \mid \text{Parents}(V_i)\} . 
\end{equation}

Using the chain rule of Bayesian Networks, the joint probability distribution over all nodes \${V\_1, V\_2, \ldots, V\_n}\$ is given by~\cite{villani2008optimal}:

\begin{equation}
\Pr\{V_1, V_2, \ldots, V_n\} = \prod_{i=1}^{n} \Pr\{V_i \mid \text{Parents}(V_i)\}
\label{eq:joint_bn}
\end{equation}
This formulation enables the BN to infer the most probable system state by combining the semantic output of the ML model with the statistical reliability estimates from SafeML. Notably, even when contextual values (e.g., speed and distance) alone may not indicate a high-risk condition, the presence of an OOD  SafeML flag increases the posterior probability of a fallback safety mode or alert state. In our implementation,  \texttt{MLDecision} and \texttt{SafeML\_Status} are treated as observed evidence nodes. Their values condition the posterior inference in the BN, allowing for soft fusion of semantic correctness and statistical trustworthiness.

\section{Experimental Setup}

\subsection{Platooning Scenario}
The experimental scenario builds upon a platooning system proposed in [13] and extended in [10], with key enhancements that integrate ML-based perception and statistical runtime validation.  Modern platooning systems allow multiple AVs to follow a human-operated lead vehicle while maintaining a defined inter-vehicle distance. This cooperative formation, commonly referred to as Cooperative Adaptive Cruise Control (CACC), improves fuel efficiency, reduces traffic congestion, and increases overall road throughput. Each follower vehicle dynamically adjusts its speed in real time to ensure safe spacing.  Despite its operational benefits, platooning introduces safety challenges due to the close proximity between vehicles. Sudden changes in speed or unexpected environmental conditions can lead to increased collision risk. Traditional systems rely on speed limits and static parameters, which may be insufficient under real-world, dynamic conditions. A key goal is therefore to maintain safe vehicle spacing while complying with traffic regulations.
While static verification methods are commonly applied during design time, they lack adaptability at runtime. When encountering inputs that deviate from the trained distribution, the system must be able to detect such anomalies and respond accordingly. Runtime anomaly detection is thus essential for triggering safety-preserving actions in unfamiliar conditions.

\subsection{Node Design and Safety Integration}

The BN that dynamically controls safety in our platform integrates contextual and system-level indicators to infer the current safety state. Each node represents either a measurable input, a logical condition, or a latent assessment. Key nodes and their interactions are described below:

\textbf{ML Decision (Classifier Output)}: Encodes the predicted traffic sign class (e.g., speed limits) from a CNN trained on the GTSRB dataset. The predicted class is used as a semantic input to inform downstream nodes about the expected driving behaviour.

\textbf{SafeML Status:} Receives the statistical decision from SafeML, which estimates whether an input sample is within the training ID or OOD.

\textbf{Speed Limit:} Converts the predicted traffic sign class to a numeric speed value (e.g., "30 km/h sign" class → 30 km/h).

\textbf{Speed Within Limit:} Evaluates whether the vehicle's current speed (measured via onboard sensors) is within ML-inferred legal limit.

\textbf{Speed Check:} Combines \textit{SafeML Status} and \textit{Speed Within Limit} to assess whether the traffic sign is both reliable and obeyed. It acts as a gatekeeper that links model validity and behavioral compliance.

\textbf{Safe Distance:} Determines whether the inter-vehicle distance is safe, using LIDAR and telemetry data. The outcome is binary (safe/unsafe).

\textbf{Detection Quality:} Evaluates sensor reliability, including agreement between leader and follower vehicles, and the accuracy of distance measurements. High-quality detection improves confidence in distance-based decisions.

\textbf{Is It Safe?:} A fused node that integrates outputs of \textit{Speed Check}, \textit{Safe Distance}, and \textit{Detection Quality} to determine immediate operational risk.

\textbf{{System States ($S_0$ to $S_5$):}} This final system state node aggregates multi-source observations—including model confidence, speed compliance, inter-vehicle distance, and sensor reliability—into a discrete system state that gives the current safety level. Table~\ref{tab:system_states} presents the semantics of each state based on MN inference.

\textbf{Other Supporting Nodes:} Several auxiliary nodes (e.g., \textit{Compare}, \textit{Comp\-areThreshold}, and \textit{DistanceDeviation}) are used to enable logical reasoning such as threshold checks and distance consistency, but do not directly appear in the final safety decision node.
\begin{table}[ht]
\centering
\vspace{-1em}

\caption{System states with corresponding safety conditions and recommended actions inspired by \cite{kabir2019runtime}.}
\vspace{-1em}

\label{tab:system_states}

\resizebox{\textwidth}{!}{%
\begin{tabular}{|c|p{2.5cm}|p{10cm}|}
\hline \hline
\textbf{State}\rule{0pt}{6pt} & \textbf{Name} & \textbf{Description and Recommended Action} \\ 
\hline
$S_0$ & {Fully Safe} & All safety conditions satisfied (speed, distance, SafeML, perception); proceed with normal operation. \\
\hline
$S_1$ & {\makecell[l]{Safe with \\ Uncertainty}} & System largely safe, but one component uncertain (e.g., speed or detection); continue with caution and monitor. \\
\hline
$S_2$ & {Warning} & Minor safety deviation detected (e.g., slight distance or sensor issue); drive cautiously. \\
\hline
$S_3$ & {Elevated Risk} & Major safety violations (e.g., unsafe distance or speed); immediate actions like deceleration required. \\
\hline
$S_4$ & {High Risk} & Multiple critical issues (e.g., close proximity, unreliable detection); emergency actions (e.g., hard braking) and fallback mode activation needed. \\
\hline
$S_5$ & {\makecell[l]{Critical ML \\ Failure}} & ML perception deemed unreliable (due to OOD or adversarial input); activate fallback safety mode (e.g., degraded ACC). \\
\hline \hline
\end{tabular}
\vspace{-1em}

}
\vspace{-1.5em}

\end{table}

\vspace{-1em}

\subsection{Implementation}
The dynamic safety monitoring system was implemented in Python using standard deep learning and scientific libraries. A custom Sequential CNN was trained on the German Traffic Sign Recognition. The architecture includes three convolutional layers with ReLU activation, two max-pooling layers, three dropout layers for regularization, a dense layer with 256 units (ReLU), and a final softmax layer for classification across 43 classes..Runtime reliability is assessed via a SafeML-inspired approach that uses distribution-based distance metrics to identify OOD inputs. Safety reasoning is performed with a BN implemented using standard probabilistic inference libraries. 

ML-based predictions and SafeML’s validation were integrated into the BN safety model as follows: the CNN prediction of the traffic sign (Speed Limit) is passed to the Speed Limit node. The SafeML node receives the p-value resulting from the bootstrap-based statistical test. If this p-value falls below a predefined threshold (e.g., 0.01), the system considers the input as OOD and degrades safely by transitioning into ACC mode. This hybrid BN-based structure enables the system to manage ML uncertainty in a principled way and dynamically select the safest action under varying degrees of input quality. 

Experiments were conducted using the  GTSRB dataset, a widely adopted benchmark for traffic sign classification. It includes over 50,000 color images across 43 categories of regulatory, warning, and speed signs. Images vary significantly in resolution, illumination, angle, and environmental background, making it a suitable benchmark for testing the real-world ML testing~\cite{stallkamp2011gtsrb,gtsrbkaggle2021}. For this study, 80\% of the data was used for training the CNN, and the remaining 20\% was used for validation.  The training set was typically used to learn traffic signs, while the validation and test set was utilized to assess the model's performance and reliability during deployment.  We focused on misclassified samples from the test set.  These samples were treated as potential OOD instances for the purposes of SafeML evaluation. During runtime, each test image was compared against the training samples of the predicted class rather than the ground truth class. This approach reflects a realistic deployment scenario in which only the model’s prediction is available at the time of decision-making. By using this strategy, we assessed whether SafeML can detect internal failures of the ML model, particularly those arising from distributional shifts that may not be apparent under conventional accuracy-based metrics.

\section{Results}

\subsection{SafeML Analysis of a Misclassified Sample}

To assess the robustness of the classifier, we applied \textit{SafeML} to the images that were misclassified by the CNN. This experiment investigated whether the misclassifications stemmed out of OOD inputs rather than general prediction confidence. Following SafeML, we computed the Wasserstein distance and bootstrap-based p-values between each misclassified and the ID training samples of the predicted class across the RGB channels. The results are shown in Table~\ref{tab:safeml_ood_results_updated}. As shown, the model misclassified test samples by confusing class 4 with class 3, or class 5 with class 8.  Nevertheless,  SafeML successfully flagged these cases as OOD. Specifically,  p-values for all three RGB channels were below 0.01, strongly rejecting the null hypothesis of distributional similarity with over 99\%  confidence. These findings demonstrate that SafeML of detecting even subtle statistical deviations and can effectively identify unreliable predictions.
\vspace{-1em}
\begin{table}[!htbp]
\centering
\caption{SafeML results for misclassified test samples.}
\vspace{-1 em}

\label{tab:safeml_ood_results_updated}
\scriptsize
\renewcommand{\arraystretch}{0.95}
\resizebox{\textwidth}{!}{%
\begin{tabular}{|c|c|c|c|c|c|}
\hline \hline
\textbf{OOD Detection}\rule{0pt}{6pt} & \textbf{p-value} & \textbf{Wasserstein Distance} & \textbf{Channel} & \textbf{Predicted Class} & \textbf{True Class} \\
\hline
OOD & 0.0000 & 0.0005 & 0 & 3 & 4 \\
OOD & 0.0000 & 0.0005 & 1 & 3 & 4 \\
OOD & 0.0000 & 0.0006 & 2 & 3 & 4 \\
OOD & 0.0000 & 0.0004 & 0 & 8 & 5 \\
OOD & 0.0000 & 0.0004 & 1 & 8 & 5 \\
OOD & 0.0000 & 0.0004 & 2 & 8 & 5 \\
\hline \hline
\end{tabular}

}
 \vspace{-1em}

\end{table}

\begin{figure}[!htbp]

\centering
\includegraphics[width=\linewidth]{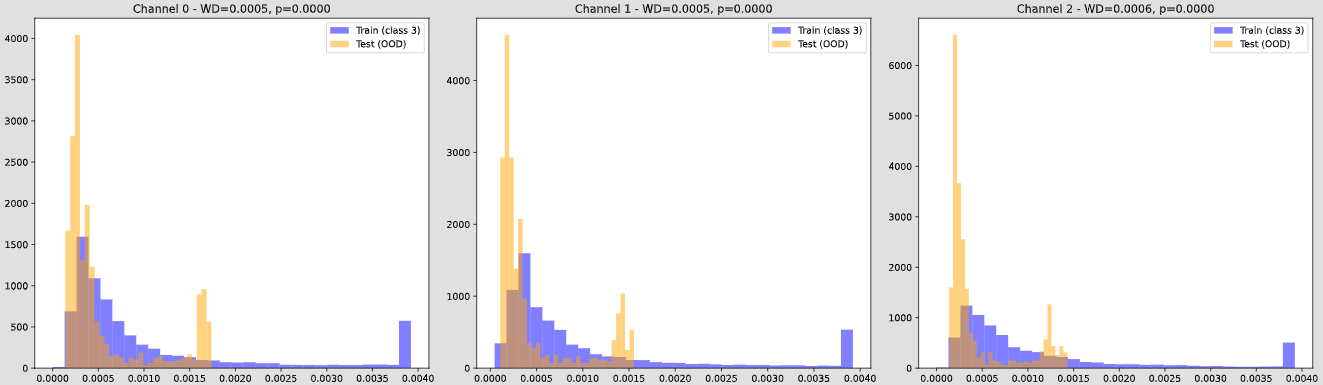}
\vspace{-2em}

\caption{Histogram of pixel values per RGB channel comparing the misclassified test image (orange) and the class 3 training distribution (blue).}
\vspace{-2em}

\label{fig:hist_channels}
\end{figure}

\begin{figure}[!htbp]
\centering

\includegraphics[width=0.7\linewidth,trim=20 15 20 0, clip]{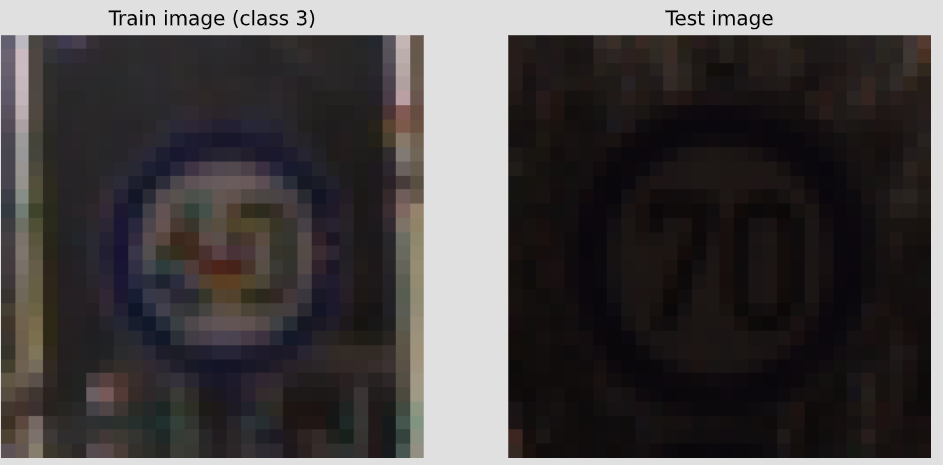}
 \vspace{-1em}

\caption{Visual comparison between a training sample from class 3 and a misclassified test image. The latter shows clear signs of darkness and low contrast.}
\label{fig:image_comparison}
 \vspace{-3em}
\end{figure}

Visual inspection further supports the statistical evidence. As shown in Figure~\ref{fig:hist_channels}, the pixel intensity histograms for each RGB channel reveal distinct deviations between the misclassified test image (orange) and the in-distribution for the predicted class (blue). The test image ---originally from class~4 but predicted as class~3--- shows a narrow, high-peaked distribution in contrast with the broader distribution of training samples.
Additionally, Figure~\ref{fig:image_comparison} compares a typical class 3 training sample (left) with the misclassified test image (right). The test image exhibits strong darkness and occlusion, confirming its deviation from the expected class distribution.
\vspace{-1em}

\vspace{-2em}

\subsection{Bayesian Inference Results and System State Evaluation}

\begin{figure}[!htbp]
\centering
\includegraphics[width=\linewidth,trim=0 20 0 0]{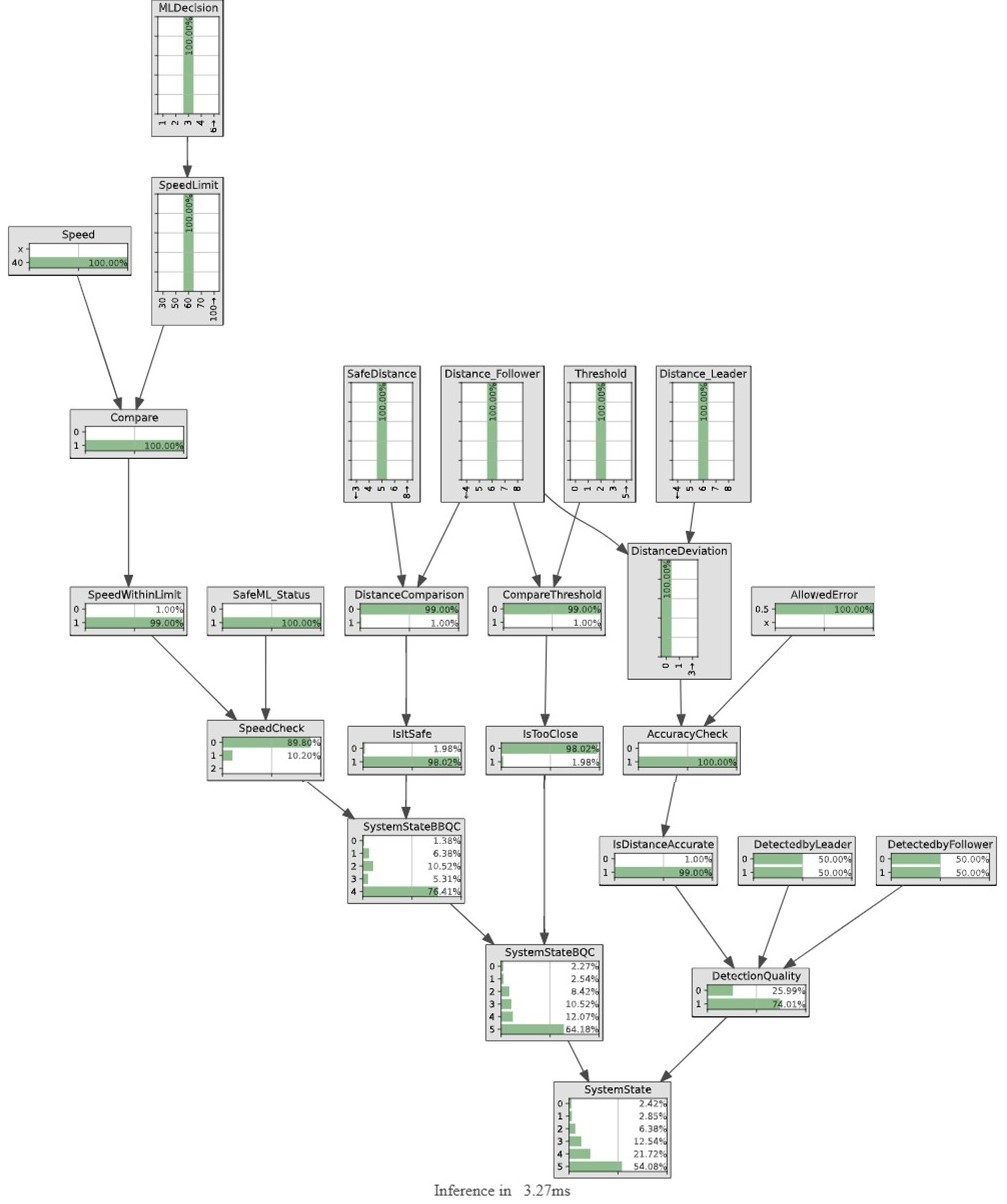}
\caption{Posterior probabilities of system states under combined SafeML and MLDecision evidence.}
\vspace{-1em}

\label{fig:bayesnet}
\end{figure}

Figure~\ref{fig:bayesnet} presents the output of the BN, executed in real-time upon detection of an \textit{OOD} input by the SafeML module. The BN integrates inputs from speed compliance, inter-vehicle distance, ML predictions, SafeML reliability checks, and sensor-level accuracy to estimate the system's most probable state.

\paragraph{\textbf{System Interpretation Based on Posterior Distribution.}}
The most likely system state is $S_5$ with a posterior probability of 54.08\%, indicating a high level of uncertainty due to unreliable sensor input or statistical anomalies. Crucially, this result demonstrates the significant influence of the SafeML node on downstream reasoning. Although speed and distance readings could not on their own justify entering a critical state, the OOD flag raised by SafeML prompts the system to transition to fallback control (ACC mode) as a precaution.

\paragraph{\textbf{Contextual Interpretation.}} Interestingly, in the evaluated scenario, contextual nodes such as \texttt{SpeedWithinLimit}, \texttt{SafeDistance}, and \texttt{DetectionQuality} remained within acceptable ranges, contributing little risk to the overall system state. Despite their nominal status, the BN still assigned the highest probability to the critical fallback state (\texttt{S5}). This highlights the dominance of the \texttt{SafeML\_Status} node, whose OOD flag alone was sufficient to override otherwise normal sensor inputs. This underscores the value of explicitly treating reasoning failure in safety assessment.
\vspace{-2em}

\vspace{-2em}

\subsection{Discussion}

To evaluate the benefit of explicit representation of ML reasoning failure in a dynamic probabilistic safety assessment, we compared the system behaviors in scenarios with and without the SafeML component. This evaluation focused specifically on the system's ability to handle ML misclassifications and distributional shifts. The experiments were conducted under stable environmental parameters (Safe Distance = 5, Distance Follower = 6, Threshold = 2, Distance Leader = 6, Allowed Error = 0.5). Detailed numerical results are presented in Tables~\ref{tab:without_safeml} and~\ref{tab:fn_low_distance}.

\subsubsection{System Performance Without SafeML}

System performance was first assessed without SafeML, relying solely on ML-based traffic sign classification outputs. Table~\ref{tab:without_safeml} summarizes the results under different scenarios.

\begin{table}
\vspace{-1em}

\centering
\caption{System performance without SafeML under various ML misclassification conditions.}
\vspace{-0.5em}

\label{tab:without_safeml}

\scriptsize 
\renewcommand{\arraystretch}{0.95} 
\resizebox{\textwidth}{!}{%
\begin{tabular}{|l|c|c|c|c|c|c|c|c|c|c|}
\hline \hline
{No.}\rule{0pt}{6pt} & {MLDecision} & \makecell{True Class} & \makecell{Speed Limit} & {Speed} & {S0} & {S1} & f{S2} & {S3} & {S4} & {S5} \\
\hline
1 & Speed limit > Speed & 5 & 80 & 60 & \textbf{0.4247} & 0.1372 & 0.1169 & 0.1513 & 0.1293 & 0.0407 \\

2 &Speed limit < Speed & 3 & 60 & 70 & 0.1019 & 0.0900 & 0.2049 & \textbf{0.3179} & 0.2456 & 0.0397 \\

3 & Speed limit = Speed & 5 & 80 & 80 &\textbf{0.4247} & 0.1372 & 0.1169 & 0.1513 & 0.1293 & 0.0407 \\

4 & Speed limit > Speed & 5 & 30 & 80 & \textbf{0.4247} & 0.1372 & 0.1169 & 0.1513 & 0.1293 & 0.0407 \\

5 & Speed limit < Speed & 2 & 30 & 60 & 0.1019 & 0.0900 & 0.2049 & \textbf{0.3179} & 0.2456 & 0.0397 \\

6 & Speed limit = Speed & 3 & 60 & 60 & \textbf{0.4247} & 0.1372 & 0.1169 & 0.1513 & 0.1293 & 0.0407 \\
\hline \hline
\end{tabular}
}
\vspace{-1em}

\end{table}
\vspace{-0.5em}

In scenarios where the ML classifier correctly identified the traffic sign, the systems reliably inferred a safe state. However, in cases of misclassification, particularly when the actual speed exceeded the correct speed limit, the system continued to report a fully safe state. This outcome reveals overreliance on potentially dangerous incorrect ML outputs. Rows 1 to 3 show correct classifications, whereas rows 4 to 6 represent misclassification cases.
\vspace{-0.5em}


\subsubsection{Impact of SafeML Integration}
In a second set of experiments, we used SafeML to distinguish between ID and OOD inputs. This integration allowed the system to dynamically adjust its behavior based on information about whether to trust or not traffic sign recognition and informed more nuanced transitions between safety states in the BN. Table~\ref{tab:fn_low_distance} illustrates the impact of SafeML intervention on these transitions under various contextual conditions.

The impact of SafeML is evident in rows 1-8 where \texttt{SafeML\_Status = 1}. In each of these cases,  the system assigns the highest probability to state \textbf{S5}, signaling a shift to the most conservative safety response. Notably, this includes instances like rows 5 and 6, where the ML prediction was correct, but SafeML incorrectly flagged the input as ODD. Despite the false alarm, the system's fallback to a safe state is non-disruptive and aligns with safety-first design priorities. In contrast, rows 9 and 10 illustrate the system’s behavior when SafeML did not intervene (\texttt{SafeML\_Status = 0}). Here, the safety outcome is dictated solely by the ML prediction and contextual parameters such as speed. In row 9, the speed exceeds the (misclassified) speed limit, causing the system to activate state \textbf{S3}, indicating elevated risk. In row 10, the speed is below the predicted limit, resulting in assignment to state \textbf{S0}—considered fully safe. These outcomes highlight the limitations of relying on ML outputs alone: erroneous inferences can propagate unchecked in the absence of runtime validation.

\begin{table}
\vspace{-1em}

\centering
\caption{System state probabilities for misclassified samples under low statistical distance.}
\vspace{-0.5em}

\label{tab:fn_low_distance}
\scriptsize
\renewcommand{\arraystretch}{0.95}
\resizebox{\textwidth}{!}{%
\begin{tabular}{|l|c|c|c|c|c|c|c|c|c|c|c|}
\hline \hline
{No.}\rule{0pt}{6pt} & {SafeML\_Status} & {MLDecision} & {True Class} & {Speed Limit} & {Speed} & {S0} & {S1} & {S2} & {S3} & {S4} & {S5} \\
\hline
1 & 1 & 3 & 1 & 60  & 40  & 0.0242 & 0.0285 & 0.0638 & 0.1254 & 0.2172 & \textbf{0.5408} \\
2 & 1 & 3 & 8 & 60  & 90  & 0.0302 & 0.0410 & 0.0997 & 0.1761 & 0.2281 & \textbf{0.4249} \\
3 & 1 & 3 & 7 & 60  & 40  & 0.0242 & 0.0285 & 0.0638 & 0.1254 & 0.2172 & \textbf{0.5408} \\
4 & 1 & 5 & 1 & 80  & 40  & 0.0242 & 0.0285 & 0.0638 & 0.1254 & 0.2172 & \textbf{0.5408} \\
5 & 1 & 5 & 5 & 80  & 90  & 0.0302 & 0.0410 & 0.0997 & 0.1761 & 0.2281 & \textbf{0.4249} \\
6 & 1 & 5 & 5 & 80  & 40  & 0.0242 & 0.0285 & 0.0638 & 0.1254 & 0.2172 & \textbf{0.5408} \\
7 & 1 & 8 & 5 & 120 & 40  & 0.0242 & 0.0285 & 0.0638 & 0.1254 & 0.2172 & \textbf{0.5408} \\
8 & 1 & 3 & 4 & 60  & 40  & 0.0242 & 0.0285 & 0.0638 & 0.1254 & 0.2172 & \textbf{0.5408} \\
9 & 0 & 3 & 3 & 60  & 90  & 0.1019 & 0.0900 & 0.2049 & \textbf{0.3179}  & 0.2456 & 0.0397 \\
10 & 0 & 4 & 4 & 70  & 40  & \textbf{0.4247} & 0.1372 & 0.1169 & 0.1513 & 0.1293 & 0.0407 \\
\hline \hline
\end{tabular}
}
\vspace{-1.5em}

\end{table}
\vspace{-1.5em}


\section{Conclusion}

This paper introduced a probabilistic runtime safety assurance framework that combines statistical reliability assessment using SafeML with semantic decision-making via Bayesian Networks. The approach explicitly models ML failures, particularly those arising from OOD inputs, and integrates them into system-level reasoning for adaptive control in autonomous vehicle platooning. Experimental results demonstrated that SafeML significantly improves the system’s ability to detect and mitigate high-risk scenarios, especially when ML predictions are unreliable. Crucially, the integration allowed the system to transition into conservative fallback states in response to statistical deviations, even in cases where conventional indicators (e.g., speed or distance compliance) suggested no apparent risk. This underscores the importance of incorporating distributional awareness into safety-critical ML applications.

While the framework exhibits strong performance in capturing latent hazards, occasional false negatives from SafeML indicate the need for enhancement. Nevertheless, the system maintained a robust and conservative response profile, consistent with safety-by-design principles. Future work will extend the framework to support temporal reasoning and dynamic sensitivity adjustment of SafeML under varying uncertainty levels. Additionally, we plan to integrate dynamic context-aware fallback strategies and countermeasure selection into the BN to support proactive safety management. Finally, validation in more complex, multi-agent scenarios and real hardware will be pursued to evaluate real-time performance and real-world applicability.




\bibliographystyle{splncs04}
\bibliography{references}

\begin{thebibliography}{10}
\providecommand{\url}[1]{\texttt{#1}}
\providecommand{\urlprefix}{URL }
\providecommand{\doi}[1]{https://doi.org/#1}

\bibitem{akram2022stadre}
Akram, M.N., et~al.: Stadre and stadro: Reliability and robustness of ml forecasting using s-d measures. In: SAFECOMP 2022 Wkps. pp. 289--301. Springer (2022)

\bibitem{aslansefat2021toward}
Aslansefat, K., Kabir, S., Abdullatif, A., Vasudevan, V., Papadopoulos, Y.: Toward improving confidence in autonomous vehicle software: A study on traffic sign recognition systems. Computer  \textbf{54}(8),  66--76 (2021)

\bibitem{aslansefat2020safeml}
Aslansefat, K., Sorokos, I., Whiting, D., Tavakoli~Kolagari, R., Papadopoulos, Y.: Safeml: safety monitoring of machine learning classifiers through statistical difference measures. In: IMBSA 2020, Lisbon. pp. 197--211. Springer (2020)

\bibitem{aslansefat2022safedrones}
Aslansefat, K., et~al.: Safedrones: Real-time reliability evaluation of uavs using eddis. In: IMBSA 2022, Munich. pp. 252--266. Springer (2022)

\bibitem{aslansefat2023explaining}
Aslansefat, K., et~al.: Explaining black boxes with a smile: Statistical model-agnostic interpretability with local explanations. IEEE Software  (2023)

\bibitem{cho2022online}
Cho, H., Lee, K., Choi, N., Kim: Online safety zone estimation and violation detection for nonstationary objects in workplaces. IEEE Access  \textbf{10},  39769--39781 (2022)

\bibitem{DIN92005}
{DIN SPEC 92005: Machine Learning – Uncertainty Quantification}. Tech. rep., Berlin, Germany (2022), \url{https://www.din.de/en/wdc-beuth:din21:343195966}

\bibitem{fagnant2015preparing}
Fagnant, D.J., Kockelman, K.: Preparing a nation for autonomous vehicles. Transport Research Part A: Policy and Practice  \textbf{77},  167--181 (2015)

\bibitem{farhad2022keep}
Farhad, et~al.: Keep your distance: Determining sampling and distance thresholds in ml monitoring. In: IMBSA 2022, Munich. pp. 219--234. Springer (2022)

\bibitem{farhad2023scope}
Farhad, A.H., Sorokos, I., Akram, M.N., Aslansefat, K., Schneider, D.: Scope compliance uncertainty estimate. arXiv preprint arXiv:2312.10801  (2023)

\bibitem{gautam2021runtime}
Gautam, V., Gheraibia, Y., Alexander, R., Hawkins, R.D.: Runtime decision making under uncertainty in autonomous vehicles. In: Proceedings of the Workshop on Artificial Intelligence Safety (SafeAI 2021). CEUR Workshop Proceedings (2021)

\bibitem{kabir2019applications}
Kabir, S., Papadopoulos, Y.: Applications of bns and petri nets in safety, reliability, and risk assessments: A review. Safety science  \textbf{115},  154--175 (2019)

\bibitem{kabir2019runtime}
Kabir, S., Sorokos, I., Aslansefat, K., Papadopoulos, Y., Gheraibia, Y., Reich, J., Saimler, M., Wei, R.: A runtime safety analysis concept for open adaptive systems. In: IMBSA 2019, Thessaloniki. pp. 332--346. Springer (2019)

\bibitem{kabir2018dynamic}
Kabir, S., Walker, M., Papadopoulos, Y.: Dynamic system safety analysis in hip-hops with petri nets and bayesian networks. Safety science  \textbf{105},  55--70 (2018)

\bibitem{kabir2022combining}
Kabir, S., et~al.: Combining drone-based monitoring and machine learning for online reliability evaluation of wind turbines. In: Intl Conf.on Computing, Electronics \& Communications Engineering (iCCECE). pp. 53--58. IEEE (2022)

\bibitem{gtsrbkaggle2021}
{Kaggle Contributor}: {GTSRB - German Traffic Sign Recognition Benchmark}. \url{https://www.kaggle.com/datasets/meowmeowmeowmeowmeow/gtsrb-german-traffic-sign} (2021)

\bibitem{mcallister2017concrete}
McAllister, R.T., Gal, Y., Kendall, A., Van Der~Wilk, M., Shah, A., Cipolla, R., Weller, A.: Concrete problems for autonomous vehicle safety: Advantages of bayesian deep learning. Int'l Joint Conferences on AI, Inc. (2017)

\bibitem{muller2016safety}
M{\"u}ller, S., Liggesmeyer, P.: Safety assurance for emergent collaboration of open distributed systems. In: ISSREW. pp. 249--256. IEEE (2016)

\bibitem{reich2016systematic}
Reich, J.: Systematic engineering of safe open adaptive systems shown for truck platooning. Ph.D. thesis, TU Kaiserslautern (2016)

\bibitem{schneider2013conditional}
Schneider, D., Trapp, M.: Conditional safety certification of open adaptive systems. ACM Trans Autonomous and Adaptive Systems (TAAS)  \textbf{8}(2),  1--20 (2013)

\bibitem{stallkamp2011gtsrb}
Stallkamp, J., et~al.: The german traffic sign recognition benchmark: A multi-class classification competition. In: IJCNN. pp. 1453--1460. IEEE (2011)

\bibitem{tsugawa2016review}
Tsugawa, S., Jeschke, S., Shladover, S.E.: A review of truck platooning projects for energy savings. IEEE Transactions on Intelligent Vehicles  \textbf{1}(1),  68--77 (2016)

\bibitem{villani2008optimal}
Villani, C.: Optimal Transport: Old and New, Grundlehren der Mathematischen Wissenschaften, vol.~338. Springer, Berlin (2008)

\end{thebibliography}

\end{document}